# Artificial Intelligence and the Spatial Documentation of Languages

**Hakam Ghanim**
**Department of Linguistics**
**Carleton University**
0009-0001-9370-3236
hakamghanim@cmail.carleton.ca
+1 3435965653

**Abstract**

The advancement in technology has made interdisciplinary research more accessible. Particularly, the breakthrough in Artificial Intelligence (AI) has given huge advantages to researchers working in interdisciplinary and multidisciplinary fields. This study investigates the ability of AI models, particularly GPT-4 and GPT Data Analyst, in creating language maps for language documentation. The study Integrates documentary linguistics, linguistic geography, and AI by showcasing how AI models facilitate the spatial documentation of languages through the creation of language maps with minimal cartographic expertise. The study is conducted using a CSV file and a GeoJSON file both obtained from HDX and from the researcher's fieldwork. The study data is then applied in real-time conversations with the AI models in order to generate the language distribution maps. The study highlights the two AI models capabilities in generating high-quality static and interactive web maps and streamlining the map-making process, despite facing challenges like inconsistencies and difficulties in adding legends. The findings suggest a promising future for AI in generating language maps and enhancing the work of documentary linguists as they collect their data in the field, pointing towards the need for further development to fully harness AI's potential in this field.

### 1- Introduction

The evolution of technology has profoundly shaped the field of language documentation, marking a journey from the humble pen and notebook to the sophisticated realms of digital mapping and artificial intelligence. Initially, linguists armed with notebooks ventured into fieldwork to document languages, a method later enhanced by the emergence of portable audio devices. These tools allowed for repeated playback of recorded speech, significantly improving the accuracy of transcriptions. As technology advanced, the introduction of microphones, laptops, and cameras further enriched this process. These devices not only captured higher quality audio but also facilitated the storage of data and documented non-verbal aspects of language through video recordings.

Furthermore, emerging software like ELAN and FLEX further revolutionized the field by enabling the recording, annotation, and analysis of linguistic data for archival in online repositories. Additionally, the era of data visualization expanded the role of maps beyond mere geographic representation, integrating them into geo-linguistics and linguistic geography as tools for visualizing language data to be used in the spatial documentation of languages. This paper explores the potential of an interdisciplinary technology, specifically Artificial Intelligence (AI) exemplified by GPT-4 and GPT Data Analyst, to generate language maps that enhance the spatial documentation of languages and make it more accessible to documentary linguists with little expertise in cartography or geography. Subsequent sections will look into documentary linguistics, geo-linguistics, linguistic geography, and language mapping, culminating in a discussion on the application of OpenAI's Large Language Models in this context. The next section elaborates on the field of language documentation.

**1.1. Language Documentation**

Language documentation, an essential field within linguistics, arose from the urgent need to preserve the world's endangered languages. It addresses a concerning reality: out of approximately 7,000 languages (Campbell & Belew, 2018), over half face the threat of extinction. This count, which includes both spoken and signed languages, has been documented by governments, public institutions, and academic communities. Within these 7000 languages, 2,698 are endangered or unsafe, 2,362 definitely endangered, 1,163 potentially vulnerable, 463 severely endangered, and 383 critically endangered. In contrast, only 65 languages are deemed safe. Krauss (1992) alarmingly predicts that, in a worst-case scenario, 90% of these languages could be extinct within a century, while in the best-case scenario, only half of them might survive.

Himmleman (2006) defines language documentation as a field of linguistic inquiry and practice in its own right, primarily concerned with compiling and preserving linguistic primary data, and interfacing this data with various types of analyses. Similarly, Woodbury (2010) describes it as the creation, annotation, preservation, and dissemination of language data and records collected during linguistic fieldwork. This multidisciplinary field, drawing on linguistics, ethnography, psychology, computer science, and recording arts (Harrison 2005; Coelho 2005; Eisenbeiss 2005), not only conserves languages but also safeguards the scientific and cultural knowledge they embody. For instance, the loss of a language is similar to the extinction of a species, underscoring the profound impact of this field's endeavor. The following section explores the fields of language mapping and linguistic geography and their connection to language documentation.

## 1.2. Geo-linguistics, Linguistic Geography, and Language Mapping

The field of language documentation is inherently multidisciplinary, encompassing a wide array of disciplines including linguistics, sociology, computer science, biology, zoology, botany, anthropology, and geography to name a few. This paper focuses on linguistic geography and geo-linguistics within the scope of language documentation. Linguistic geography, often associated with dialectology, has a narrower scope compared to the broader reach of dialectology, as noted by Trudgill and Chambers (1998). Geo-linguistics, as outlined by Stone (2020), explores the distribution of languages and their features—referred to as micro geo-linguistics—as well as the relationship between languages and their geographical contexts. Stone (2020) asserts that both fields fundamentally align in their use of language maps to analyze the distribution and characteristics of languages and linguistic features.

The practice of documenting languages through mapping and has historical precedence. For instance, Gawne (2016) discusses the creation of language and linguistic maps as methods to spatially document languages. Furthermore, a significant number of the world's languages are under-researched and undocumented, leading to a limited understanding of their geographical distribution. Often, the precise locations where these languages are spoken remain known only to the native speakers themselves (Dahl & Veselinova, 2005).

The tradition of mapping languages in linguistic research dates back several centuries. Early examples in Western linguistic tradition include the 'Sprachatlas des Deutschen Reiches' by Georg Wenker and Ferdinand Wrede (1888), and 'Atlas Linguistique de la France' by Jules Gilliéron (1902). Since these pioneering efforts, linguists have employed maps for diverse purposes, such as historically mapping dialects, studying sociolinguistic variation, examining language attitudes and perceptions, and observing language variation and change over time, among other linguistic inquiries (Moore et al. 1935; Trudgill 1983; Labov et al. 2006; Gawne, 2016; Waller 2009; Kretzschmar 2013b; Alfaraz 2014; Kulkarni et al. 2014; MacKenzie 2014; Prichard 2014; Bounds 2015; Yan 2015).

Additionally, Canvin and Tucker (2020) have developed language maps that categorize endangered languages based on their levels of endangerment, utilizing the 'Graded Intergenerational Disruption Scale' (GIDS) and the 'Expanded Graded Intergenerational Disruption Scale' (EGIDS). The GIDS, formulated by Fishman (1990), delineates eight levels of language endangerment, whereas the EGIDS, developed by UNESCO (2003), expands upon this framework. Canvin and Tucker (2020) argue that maps which display languages within their geographical contexts and indicate their vitality and endangerment levels are invaluable tools. They aid in formulating strategies to maintain cultural and linguistic diversity, such as promoting the use of heritage languages, planning educational services, developing communication strategies, and securing recognition for these languages.

Lastly, the World Atlas of Language Structures (WALS) project, undertaken by the Department of Linguistics at the Max Planck Institute for Evolutionary Anthropology in Leipzig, Germany, spanned five years (1999–2004). The project's outcomes, initially published in print form by Haspelmath et al. (2005) and supplemented by an Interactive Reference Tool on CD-ROM developed by H.-J. Bibiko, sought to depict the geographical distribution and structural diversity of the world's languages. The WALS project focuses on mapping various levels of linguistic structure, including phonetics/phonology, syntax, and lexicon, as elucidated by Comrie (2019). The next section provides an insight to the AI models used in this study.

### 1.3. OpenAI Large Language Models (LLMs)

The field of Natural Language Processing has seen significant advancements with the introduction of Large Language Models (LLMs) such as ChatGPT developed by OpenAI. These models, characterized by their transformer-based architecture, have been trained on extensive text corpora that enables them to generate text across a variety of styles and formats. With a profound grasp of human language, LLMs like ChatGPT have been applied in diverse areas including but not limited to reasoning, educational curriculum design, creative writing, code generation, translation, image creation, data analysis, and even in geography and GIS (Xin Hua et al., 2023; Tlili, 2023; Hu, 2023). This research explores the ability of such a model in creating language maps that can contribute to the spatial documentation of languages as well as maps that show the distribution of languages. The data used in this study is a sample from a larger data set on the distribution of languages in al-Hamdaniyah district in Ninawa Province, in northern Iraq.

The organization of this paper is as follows: Section two looks into the technological tools used in language mapping. Section three outlines the research context, poses critical questions, and provides a review of literature on the integration of AI in geography and cartography. Section four describes the methodologies and data acquisition strategies employed to conduct the study. Section five shows the results of utilizing the AI models ChatGPT-4 and GPT Data Analyst to generate language maps. Section six engages in a discussion on the resulted maps, and section seven concludes the study.

### 2. Technologies in Language Mapping

Canvin and Tucker (2020) and Peeters (1992) have noted that historically, map-making was restricted to those with the necessary technical skills and financial resources, creating a disparity in representation. However, this landscape has transformed dramatically with technological advancements. The range of available map-making software is broader than ever, with many being open source, although some require payment for full functionality. Notable examples include QGIS[1], ArcGIS[2], SagaGis[3], Grass GIS[4], gvSIG[5], and OpenJUMP[6]. Furthermore, the evolution of mapping technology has led to the rise of web mapping, a process defined by Neumann (2011) as designing, implementing, generating, and delivering maps via the internet. Several web mapping platforms, such as Open Street Map[7], Google Maps[8], Mapbox[9], World Map[10], and D3js[11], have found applications in linguistic research and the creation of language maps (Stone, 2020).

In addition to these technologies, significant breakthroughs in Natural Language Processing (NLP) and the development of Large Language Models (LLMs) have revolutionized data visualization and mapping. The use of Artificial Intelligence (AI) in this domain has become more accessible and user-friendly, even for those with only basic knowledge in the field. This research paper focuses primarily on the integration of AI, GPT-4 and GPT Data Analyst, in geography and mapping as well as investigating the ability of these AI models in creating language maps.

### 3. Research Context

This research operates at the intersection of artificial intelligence, documentary linguistics, and linguistic geography, addressing the capability of AI tools, specifically ChatGPT, to generate language maps from researcher-provided data. The data used to generate the language maps focuses is on al-Hamdaniyah district in Ninawa Province, north of Iraq. The next section of this part of the paper provides a literature review on other related works and studies.

---

[1] www.qgis.org

[2] www.arcgis.com

[3] www.saga-gis.org

[4] www.grass.osgeo.org

[5] www.gvsig.com

[6] www.openjump.org

[7] https://www.openstreetmap.org/#map=16/36.3192/43.4715&layers=Y

[8] https://www.google.com/maps

[9] https://www.mapbox.com

[10] https://about.worldmap.harvard.edu

[11] https://d3js.org

## 3.1. Literature Review

This section of the research provides an overview of some previous studies on the use of Artificial Intelligence represented by ChatGPT-4 and GPT Data Analyst in the visualization of spatial data, geography, and mapping. For instance, Roberts, Luddecke, et al. (2023) investigated GPT-4 acquisition of factual geographic knowledge, and its ability to use this knowledge for interpretative reasoning. According to them, interpretative reasoning is vital for applications that involve geographic data such as geospatial analysis, supply chain management, and disaster response. Their experiments ranged from basic tasks like estimating locations and distances to more complex ones like generating travel networks. While GPT-4 exhibited strong logical reasoning and a comprehensive understanding of geography. They also argued that in the case that GPT-4 would have access to geographical data feeds, it may ultimately be able to create tools that improve travel planning and navigation. However, they also highlighted that it also displayed knowledge gaps, leading to occasional irrelevant results.

Another study is by Mooney et al. (2023) where they assessed ChatGPT's geographic proficiency by administering a GIS exam to understand its geographic capabilities and limitations, comparing the performances of GPT-3 and GPT-4. They used both GPT-3 and GPT-4 to understand whether general improvements of an LLM translate to improvements in answering questions related to the spatial domain. Their findings revealed that both models could pass an introductory GIS exam, with GPT-3 achieving a 63.3% score and GPT-4 scoring 88.3%. The assessment covered spatial analysis, mapping concepts, and data management, with both models excelling in fundamental GIS concepts. However, the limited scope of the exam questions may constrain the robustness of these findings in identifying specific areas of strength within the models.

Another important study by Tao and Xu (2023) that investigated the capabilities of ChatGPT in map-making. Providing ChatGPT with specific prompts, they utilized its responses—blocks of Python code executed in an IDE like Colab—to generate maps. ChatGPT's ability to fetch publicly available data sources and process mapping requests through conversational interactions was particularly noted. In their experiment, they prompted it to give codes to generate a thematic map of the US population as well as the generation of batch mapping. In this study, the researchers came up with the following advantages.

1- ChatGPT can significantly lower the barriers to producing maps.
2- ChatGPT can significantly boost the efficiency of map production.
3- ChatGPT can search for the right mapping functions from associated Python packages to use.
4- ChatGPT can recall prior messages within that same session for context, which simplifies the process of map improvement or extra map production via follow-up prompts.
5- ChatGPT can design mental maps based on its spatial thinking capability.
6- ChatGPT can inspire users for creative map design.

In terms of limitations, the researchers found the following limitations.

1- ChatGPT can only input and output texts.
2- ChatGPT's mapping capability is subject to external conditions.
3- ChatGPT's initial response usually does not lead to a satisfying map product.
*4-* ChatGPT benefits different users unequally.

Moreover, Li and Ning (2023) introduced the concept of Autonomous GIS, an AI-powered geographic information system designed to tackle spatial problems through automated spatial data collection, analysis, and visualization. They outlined five objectives for autonomous GIS: self-generating, self-organizing, self-verifying, self-executing, and self-growing. To illustrate this, they developed a prototype system named LLM-Geo using GPT-4's API within a Python environment, showcasing its capability to deliver accurate results, such as aggregated data, graphs, and maps, with minimal human intervention across three case studies. Despite being in preliminary stages and lacking critical components like logging and code testing, LLM-Geo demonstrated the potential for AI-powered GIS in revolutionizing spatial analysis. The researchers advocate for increased efforts within the GIS science community to advance the development of autonomous GIS systems, aiming to make spatial analysis more efficient, accessible, and user-friendly.

Finally, Kang, Zhang, and Roth (2023) highlight the ethical implications and potential risks of AI-generated maps, including those that might be produced by future Artificial General Intelligence (AGI) systems. They advocate for the development of ethical guidelines to ensure that such maps accurately represent geographic information while minimizing misinformation and bias. While Geo AI techniques have enhanced map creation processes and supported human creativity in cartographic design—evidenced by AI's role in map style transfer, retrieval, generalization, and design critique—ethical concerns have emerged. These include the uncertainty and opacity of AI-generated maps, which may lead to inaccuracies, unclear borderlines, and the inclusion of non-existent geographical entities, contributing to potential confusion and misinformation. Kang, Zhang, and Roth (2023) also examined DALL·E 2, OpenAI's advanced generative model, noting its propensity to produce maps with pseudo-words, inconsistent shapes, and truncated content due to the inherent randomness of AI generation. Such maps can be misleading, possess unanticipated features, and lack reproducibility, raising significant ethical concerns regarding their use in cartography. These observations underscore the need for careful consideration by cartographers, geographers, and GIS scientists in the development and use of AI-generated maps. Ensuring ethical and responsible usage is paramount to mitigating potential negative effects and fostering trust in AI applications within the field.

## 4. Data Collection Methodology and Data Acquisition

The primary dataset of this study comprises a CSV file and a GeoJSON file obtained from Humanitarian Data Exchange (HDX). The GeoJSON is a shape file that includes the administrative boundaries of Iraq on the district level. The GeoJSON file will be used as a base map for the static map and will also overlay the base map during the process of making the web map as will be seen later. The CSV file, initially used to map language distribution in al-Hamdaniyah district, was modified to include additional settlements not originally listed, with their coordinates sourced from Google Maps. Field visits to major settlements in al-Hamdaniyah were conducted to gather firsthand linguistic data, engaging with tribal leaders, educators, and civil activists through a simple structured questionnaire (see Appendix II). I had in mind that speakers might be biased to their language, specifically in such a linguistically and demographically diverse district as al-Hamdaniyah. Therefore, and to check for the speakers' reliability and for the sake of data verification, I interviewed more than one person from each different group of speakers of the various languages, I also did not inform any interviewee or respondent that I have met someone before them. This step was very helpful to check for the credibility of the questionnaire respondents. The data I had from all my respondents was significantly matching.

Despite the district's 148 populated places (cities, towns, and villages), hence forth referred to as settlements, direct interviews covered a representative sample, ensuring a broad understanding of the linguistic landscape. A printed reference map by the REACH Organization (2017) aided in this process, with respondents marking languages spoken across the settlements. Data enhancements included the addition of missing settlements and linguistic information to the CSV file. The linguistic information that was incorporated into the CSV file is the language distribution data showing the percentage of each of the languages in the district next to the settlements where they are spoken. Unique identification numbers for each settlement were generated using the administrative codes from HDX, reflecting Iraq's administrative hierarchy (country, province, district). The final dataset, detailed in a CSV file, includes settlement names in English and Arabic, unique IDs, administrative divisions (from province to village levels), and estimated language speaker percentages. The source of linguistic data is documented for transparency. It is worth mentioning here that the CSV file that includes the language distribution data is used in this study as a sample to aid in the investigation of the ability of the AI models GPT-4 and GPT Data Analyst in the creation of language distribution maps.

Furthermore, the approach to produce the maps using the AI models involved real-time conversations between the researcher and GPT-4 and GPT Data Analyst. The language used to ask the AI models to create the maps is simple plain English; that is, the conversations with the models did not involve any kind of technical language or the use of specific terms related to cartography, geography, or linguistics. The rationale behind this is to ascertain that these models can help any researcher with any level of knowledge of maps and cartography to produce language distribution maps. Then, the two AI models GPT-4 and GPT Data Analyst were employed and provided with data to generate language distribution maps. GPT-4 is the plus version of ChatGPT-3.5. GPT-4 requires monthly subscription to OpenAI. The subscription allows access to GPT-4 and other GPTs including GPT Data Analyst. Both GPTs used in this study allow file upload, which made it possible to upload the data files of the study and ask the models to create the maps. Additionally, they can both provide Python codes that can be copied into an external IDE, and they both provided HTML links when were asked to create web maps. However, only the GPT Data Analyst can show the resulting maps as an image, which is one feature that GPT-4 does not have.

5. Results

ChatGPT-4 and its variant GPT Data Analyst have good abilities for producing maps by providing programming codes in Python using mapping libraries such as 'Matplotlib' and 'geopandas' for producing maps as images and web maps. GPT-4 cannot directly produce maps as images; however, its variant GPT Data Analyst can produce images of the maps and present them immediately. They both understand the user's mapping request and respond with programming code blocks to fulfill the request of the user. This approach requires users to run the codes in a third-party integrated development environment (IDE) to complete the final step, it still significantly eases the entire mapping process. For this study, I adopted the free Jupyter Notebook IDE to run the codes provided by GPT-4 and GPT Data Analyst. Both GPT variants showed good results in the spatial visualization of data on static and web maps as can be seen in Appendix I. Below is a diagram that shows the steps used to produce the language maps using GPT-4 and GPT Data Analyst.

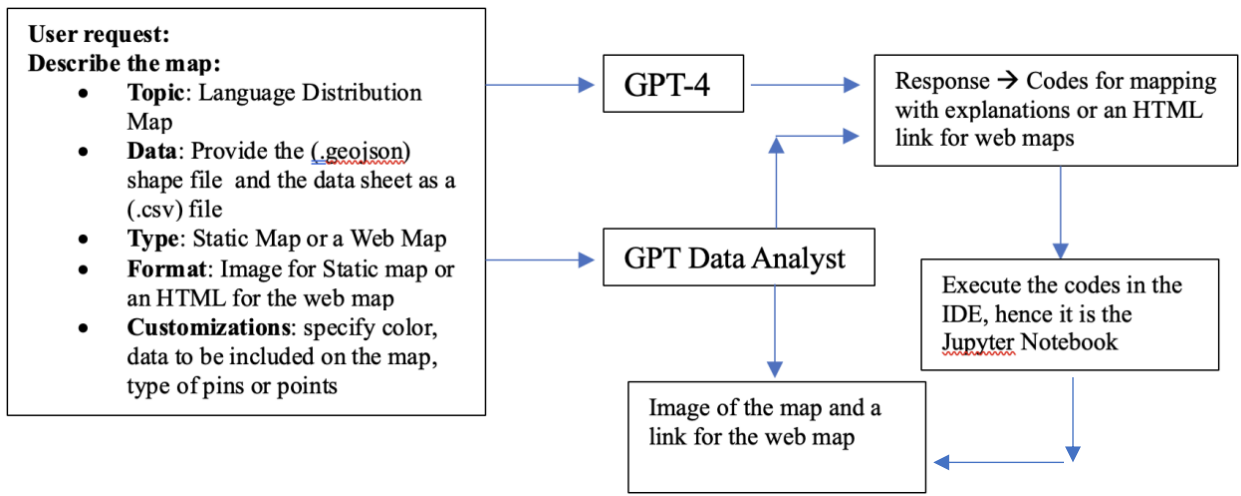

Figure (1) The process of creating language maps using GPT-4 and GPT Data Analyst

The next two sections talk about the process of creating the static maps and the web maps using both GPT variants after providing them with the (.csv) data sheet and the (.geojson) shape file.

5.1. The Static Maps

Initially, I tasked both AI models, GPT-4 and GPT Data Analyst, with creating a choropleth map to visualize language distribution, utilizing only the CSV data. The resulting code, executed in an external IDE, specifically Jupyter Notebook, produced the first series of maps, which were displayed alongside the code by GPT Data Analyst. Utilizing libraries such as pandas, geopandas, and matplotlib, the maps depicted language concentrations with color-coded points based on language prevalence in each settlement. The data's longitude and latitude were accurately represented, though the maps lacked a base layer which made them hard to understand since they only included points with different colors scattered on an empty space, as evident in Maps 1-3 in Appendix I.

Subsequently, I incorporated a GeoJSON file and asked the AI models to overlay the CSV data onto a shapefile to test the AI models' ability to incorporate the CSV data on the GeoJSON shape file. The initial step produced an administrative boundary map of Iraq (Map 4, Appendix I), the reason behind creating the administrative map of Iraq is to check for the AI models' ability to deal with the GeoJSON file. Further refinement on the administrative map of Iraq using the CSV data yielded more detailed language distribution maps showing a map of al-Hamdaniyah district with the distribution of languages spoken there. Interestingly, the AI models were able to choose the exact district I asked them to show and plot the CSV file data on it. So, the models gave enhanced visualization with varying shades of red to indicate Arabic speaker concentrations on a map of al-Hamdaniyah district that is extracted from the GeoJSON shape file. The GPTs were also able to perform batch mapping, which is the process of creating more than one maps each of which show a different aspect of the data provided using automation showcased in Map 5, Appendix I.

Finally, seeking more accurate visualizations, I requested the AI models to create maps that use point sizes to indicate language speaker percentages in each settlement. The resulting maps featured blue circles of varying sizes, with larger circles denoting higher language prevalence, as displayed in Maps 6-11 in Appendix I. being able to perform such a task means that these models possess a good ability in making maps and visualizing data using GIS data. the static maps produced clearly showed the distribution of languages in al-Hamdaniyah district.

**5.2. Web Maps**

Transitioning to interactive web maps, I initially provided the CSV data to the GPTs, requesting hover maps that reveal language distribution upon interaction. The code responses, incorporating the folium and pandas libraries, generated maps with Open Street Map as the foundational layer. These interactive maps offer zoom capabilities and detailed hover data for each settlement, enhancing user engagement. The initial web map visualized settlements as clickable pins, revealing detailed information upon interaction (Map 12, Appendix I). The codes provided by the models can be manipulated to alter the settings of the map such as choosing what to appear on the hover data upon clicking the pin.

Furthermore, exploring alternative visualizations, I prompted the creation of a point-based map, resulting in Map 13, Appendix I, which employs blue points to convey settlement data against an Open Street Map backdrop. Further experimentation with color coding to distinguish languages led to Map 14, Appendix I, where colored pins represent different languages, with clickable functionality to access embedded data.

Noticing the ambiguity in al-Hamdaniyah district borders on the Open Street Map as a base layer, I instructed the GPTs to integrate the GeoJSON shapefile with the CSV data on a web map. This approach clarified the administrative boundaries of Iraq and detailed the settlements within al-Hamdaniyah district through informative data points, as illustrated in Maps 15-16, the GeoJSON layer is overlaid on the base map showcasing the administrative boundaries of Iraq, and once zoomed in, the settlements and their hover data can be viewed upon clicking on the points representing the settlements.

**6. Discussion**

The study's findings reveal that AI models, specifically ChatGPT-4 and GPT Data Analyst, possess significant potential in cartography, particularly for creating maps tailored to linguistic research. These models demonstrated proficiency in generating both static and web maps, showcasing adaptability in handling geographic data. Notably, their ability to rectify over 95% of code errors upon feedback underscores their utility in map-making processes. Moreover, if the user has a basic knowledge in coding, they can alter the code, and eventually the resulting map by playing with the part of the code related to the style, color, title of the map, and the data to be shown on the map.

On the other hand, the study also unveiled some limitations. Despite their overall competence, the AI models exhibited inconsistencies in map production, such as variations in style and color when generating maps with corrected data. This inconsistency, particularly in the visual aspects of the maps, suggests a need for further refinement in ensuring consistent output.

Additionally, the AI models faced challenges in incorporating legends into web maps, despite code blocks intended to generate them. This limitation points to a gap in the models' ability to fully execute cartographic conventions in web-based environments.

These findings align with Tao and Xu's (2023) observations regarding AI's capacity to democratize map-making by lowering technical barriers and enhancing efficiency. This study extends their conclusions, suggesting that linguists could potentially leverage AI to integrate GIS data with linguistic fieldwork, thereby facilitating the creation of language maps with minimal cartographic expertise.

However, the limitations identified in this study, such as the absence of base maps and inconsistent outputs, mirror some of Tao and Xu's concerns, highlighting the variability in AI's utility across different use cases. Notably, this study diverges from Tao and Xu's findings regarding AI's text-only input/output limitations, possibly due to the utilization of the GPT Data Analyst Plus version, which supports file uploads and image outputs, albeit with access restrictions based on subscription plans. Moreover, in Kang, Zhang, and Roth 's (2023) study that examined DALL·E 2, OpenAI's advanced generative model, they noted that the model they used produced maps with pseudo-words, inconsistent shapes, and truncated content due to the inherent randomness of AI generation. Additionally, they claimed that such maps can be misleading, possess unanticipated features, and lack reproducibility, raising significant ethical concerns regarding their use in cartography. Their results, along with the results of this study, proves that AI models may need human help when interacting with GIS data and maps. One of the reason the maps produced in this study are more promising because all the required data such as the CSV file and the GeoJSON file were provided for the AI models.

In conclusion, while AI models like ChatGPT-4 and GPT Data Analyst provide a promising avenue for integrating cartography with linguistic research, their current limitations necessitate cautious application. Future research should focus on enhancing the consistency and completeness of AI-generated maps to ensure their reliability and comprehensibility in academic and practical contexts.

## 7. Conclusion

This study explored the application of AI models, particularly GPT-4 and GPT Data Analyst, in language documentation, focusing on the creation of language maps for spatially documenting linguistic distributions. It provided an overview of documentary linguistics and linguistic geography, setting the stage for an examination of AI's transformative role in these fields. The technology employed, notably AI, represents a significant advancement, underscoring the potential of AI in enhancing cartographic methodologies.

The findings reveal that the AI models successfully generated high-quality language distribution maps, demonstrating that linguists, even those with limited cartographic expertise, can now engage in spatial language documentation more efficiently. The interaction between the AI models and users significantly streamlined the map-making process, reducing the time and technical barriers traditionally associated with cartography.

However, the study also identified challenges, including inconsistencies in map outputs and difficulties in integrating legends into web maps. These limitations highlight areas for further development in AI's application in cartography.

The implications of this study are far-reaching, suggesting a promising future for the integration of AI in linguistic geography and documentary linguistics. It opens up new avenues for research, particularly in the development of more reliable and user-friendly AI tools for cartography. As AI continues to evolve, its potential to democratize map-making and enhance linguistic documentation could significantly contribute to the preservation and understanding of linguistic diversity.

In conclusion, while AI models like GPT-4 and GPT Data Analyst have shown considerable promise in facilitating language map creation, ongoing research and development are essential to address the current limitations and fully realize AI's potential in linguistic geography and documentation.

**Acknowledgment**
I extend my deepest gratitude to my supervisor, Erik Anonby, for his invaluable guidance, encouragement, and expert advice throughout the course of this research. His patience, knowledge, and commitment to excellence have greatly contributed to the quality of this work and have provided me with an exceptional learning experience. I am truly thankful for his unwavering support and for inspiring me to pursue my goals with dedication and rigor.

**Funding**
The authors did not receive support from any organization for the submitted work.

**Conflict of Interest**
There is no conflict of interest for this study.

Zhang, Q., Kang, Y., & Roth, R. (2023). The ethics of AI-generated maps: DALL·E 2 and AI's implications for cartography (Short Paper). In *12th International Conference on Geographic Information Science (GIScience 2023)*. Schloss Dagstuhl-Leibniz Zentrum für Informatik. https://doi.org/10.4230/LIPIcs.GIScience.2023.93

**Appendix I – Maps produced using AI models GPT-4 and GPT Data Analyst**

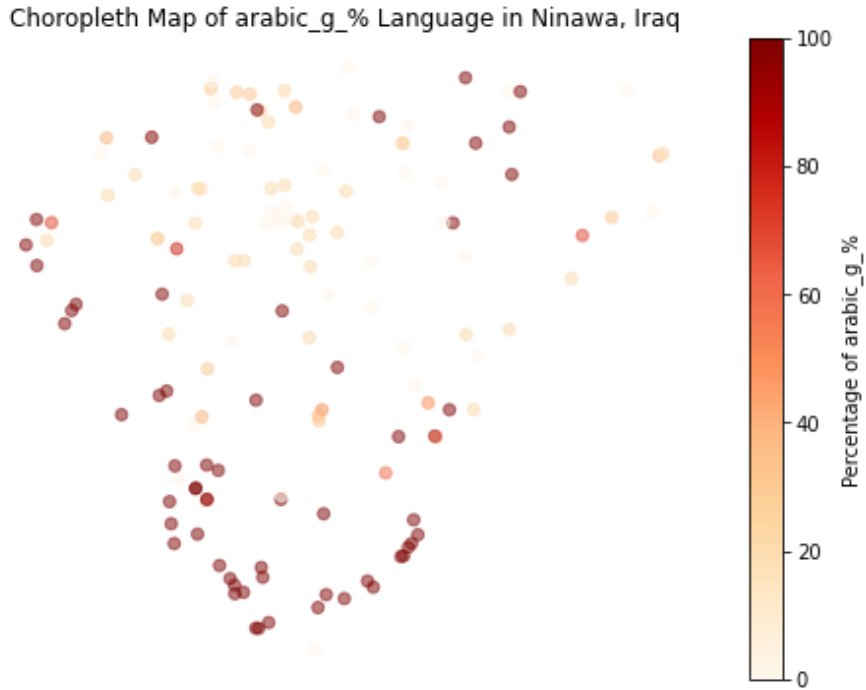

Map 1 – A choropleth map showing the distribution of Gelet Arabic in al-Hamdaniyah district.

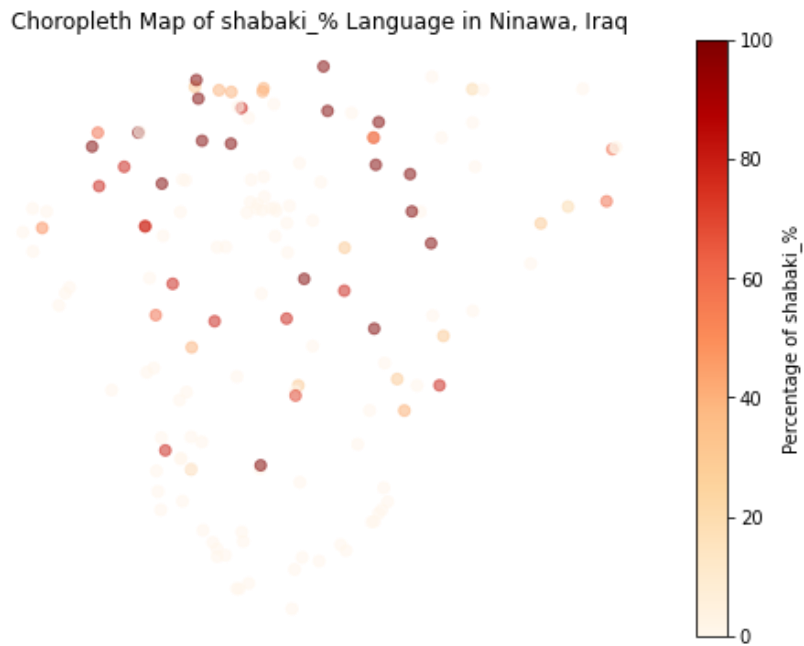

Map 2 - A choropleth map showing the distribution of Shabaki in al-Hamdaniyah district.

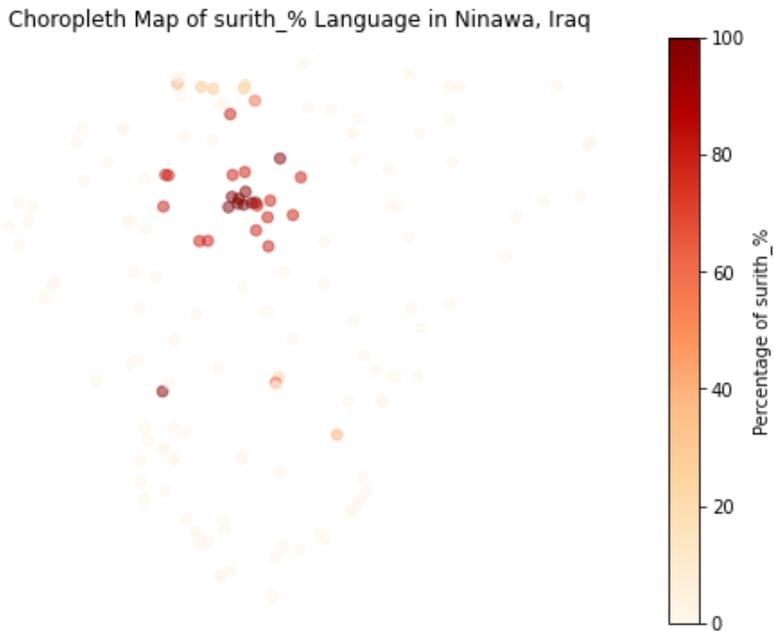

Map 3 - A choropleth map showing the distribution of Surith in al-Hamdaniyah district.

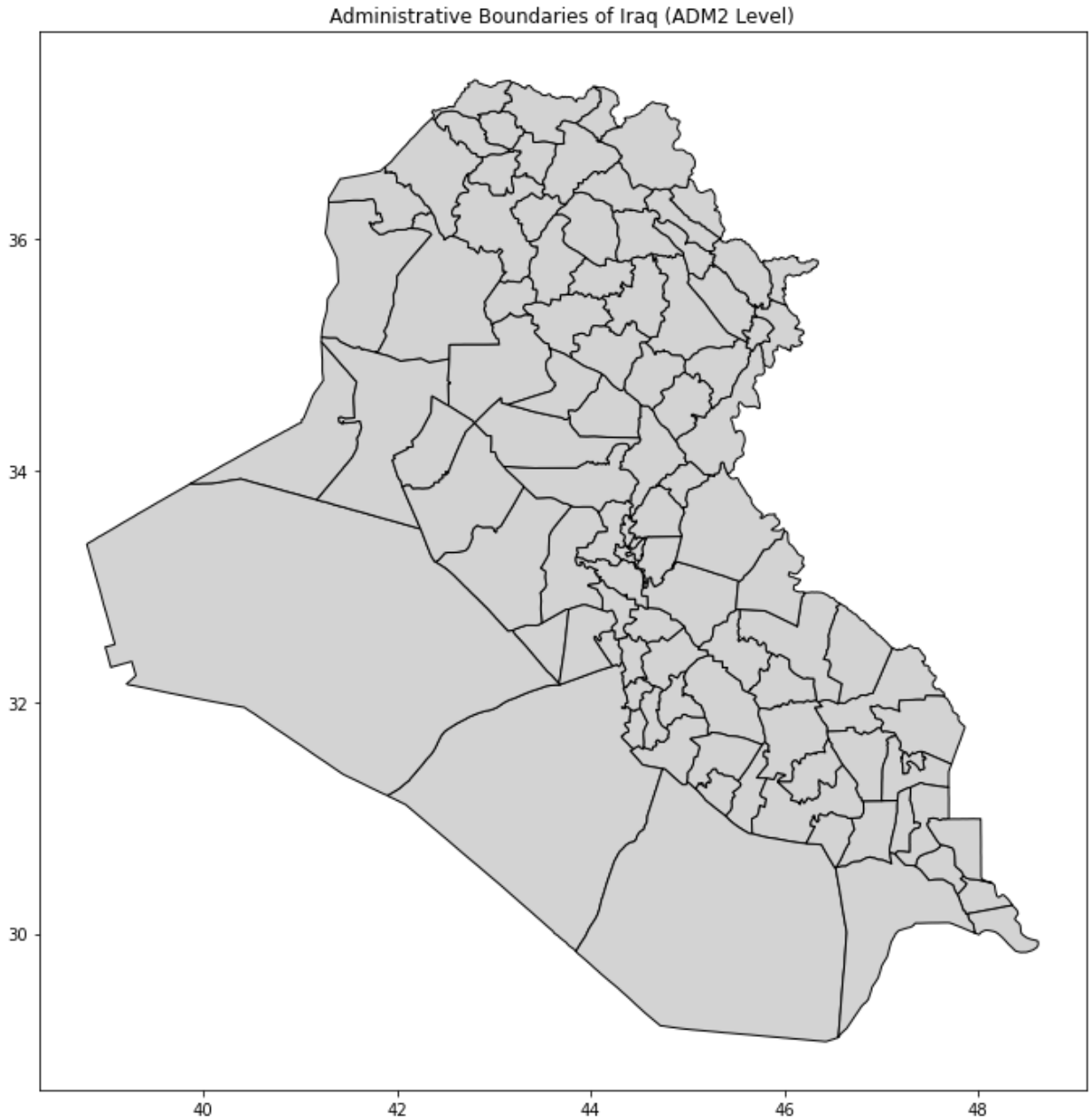

Map 4 – The administrative boundaries of Iraq after applying the models' mapping skills on the GeoJSON shape file.

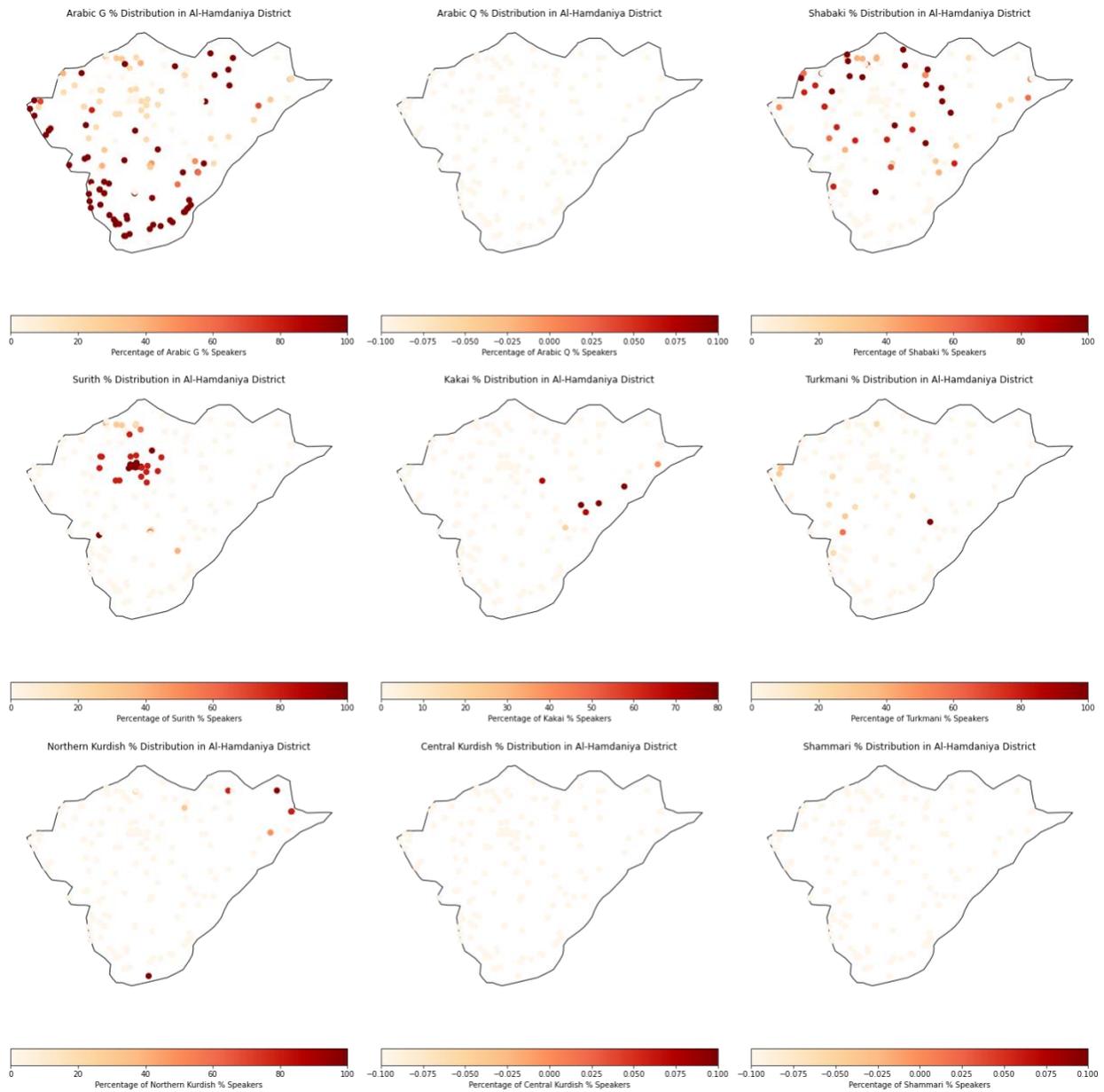

Map 5 – A group of maps that show the distribution of each of the languages in al-Hamdaniyah district. These maps were produced in a batch mapping process.

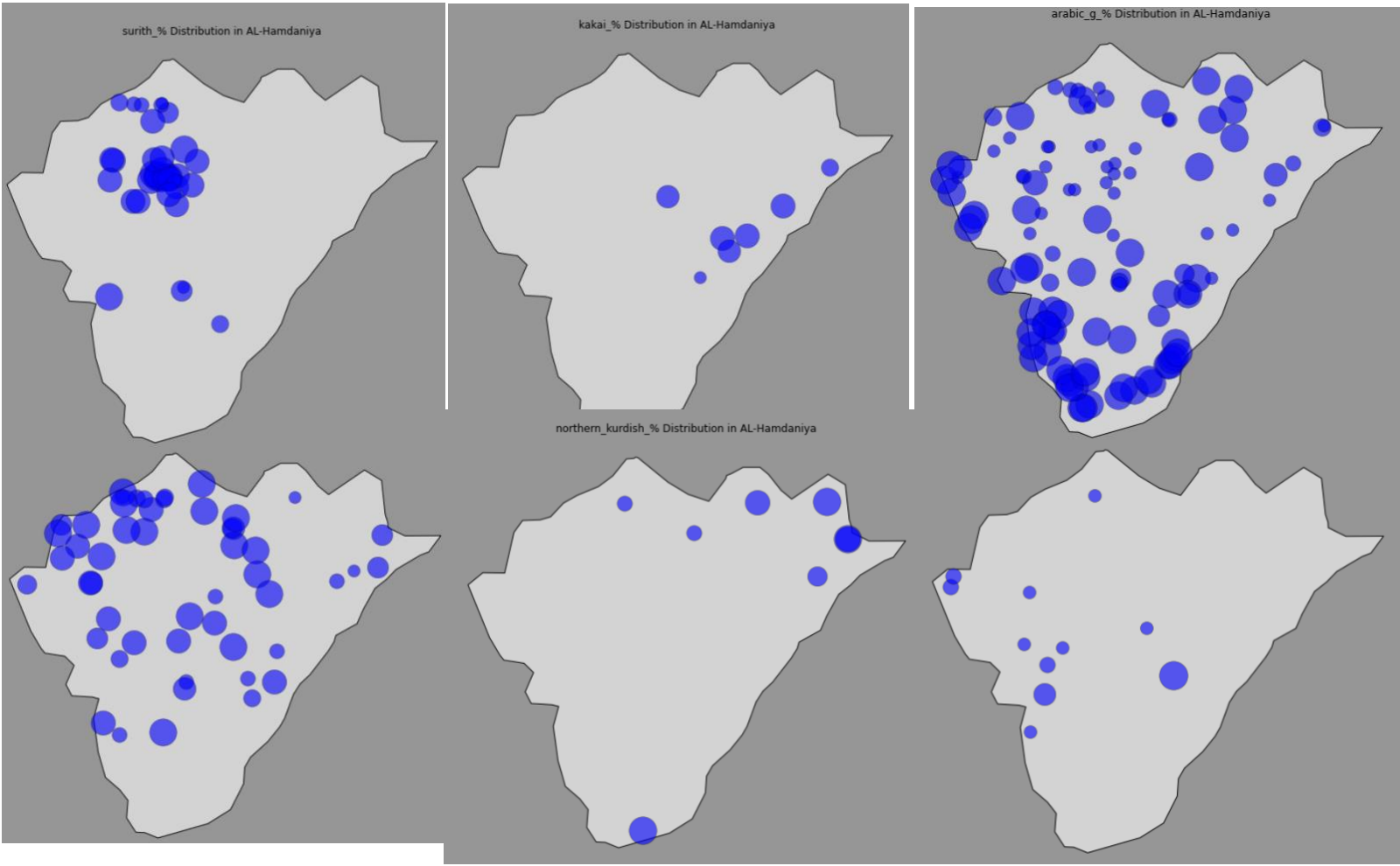

Map 6-11- These maps show the how the AI models produced maps using the map of al-hamdaniyah district from the GeoJSON shape file and applied the CSV file data on it to plot the distribution of languages in al-Hamdaniyah. The percentages of the languages are represented by the size of the circle.

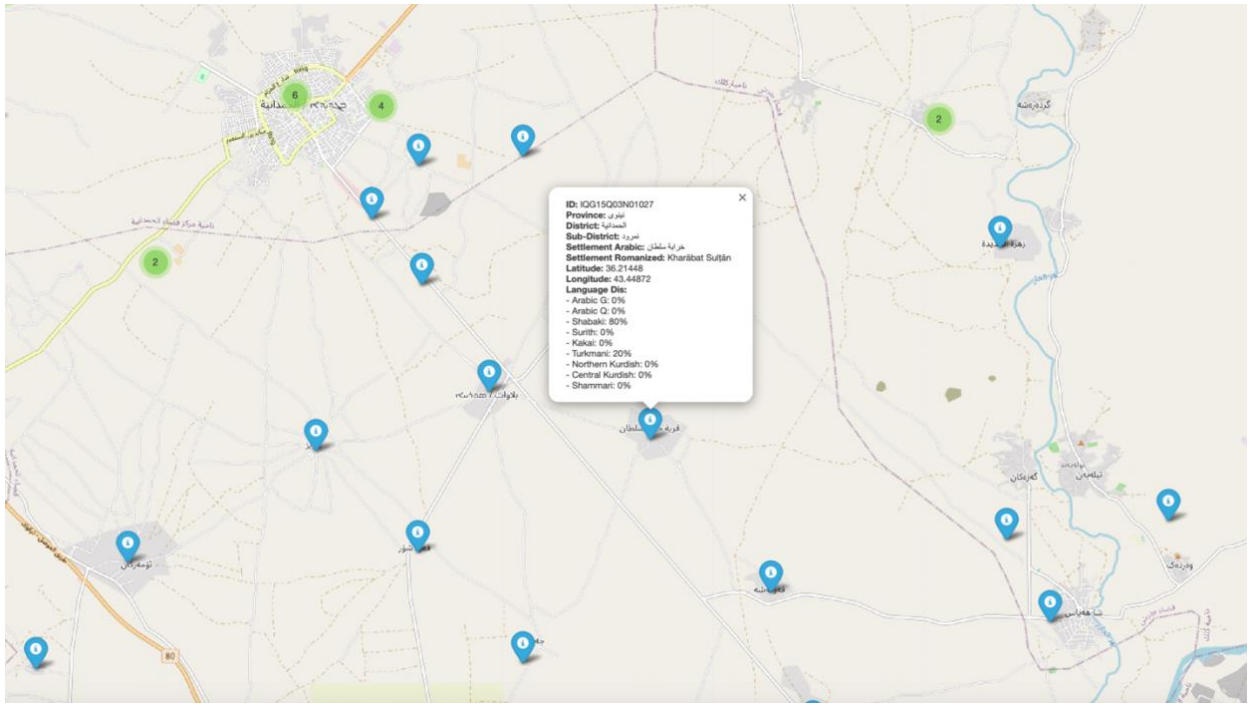

Map 12 – An interactive web map that shows the settlements of al-Hamdaniyah district and once we click on the pin, we can see all the information embedded in the hover data. Note here that we can manipulate the Python code to included whatever data we like to show.

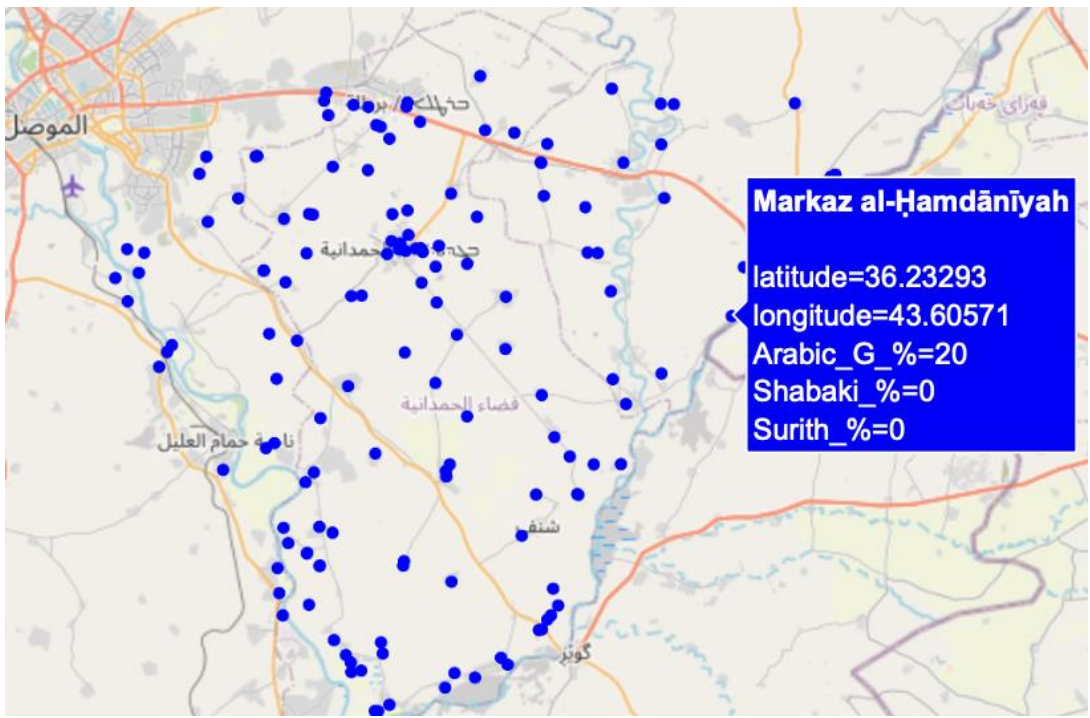

Map 13 – This is another interactive web map that shows the AI models' ability to change the way data is visualized on the map. For instance, in this map, they used points instead of pins, and once we click on these points we can see the hover data related to each settlement.

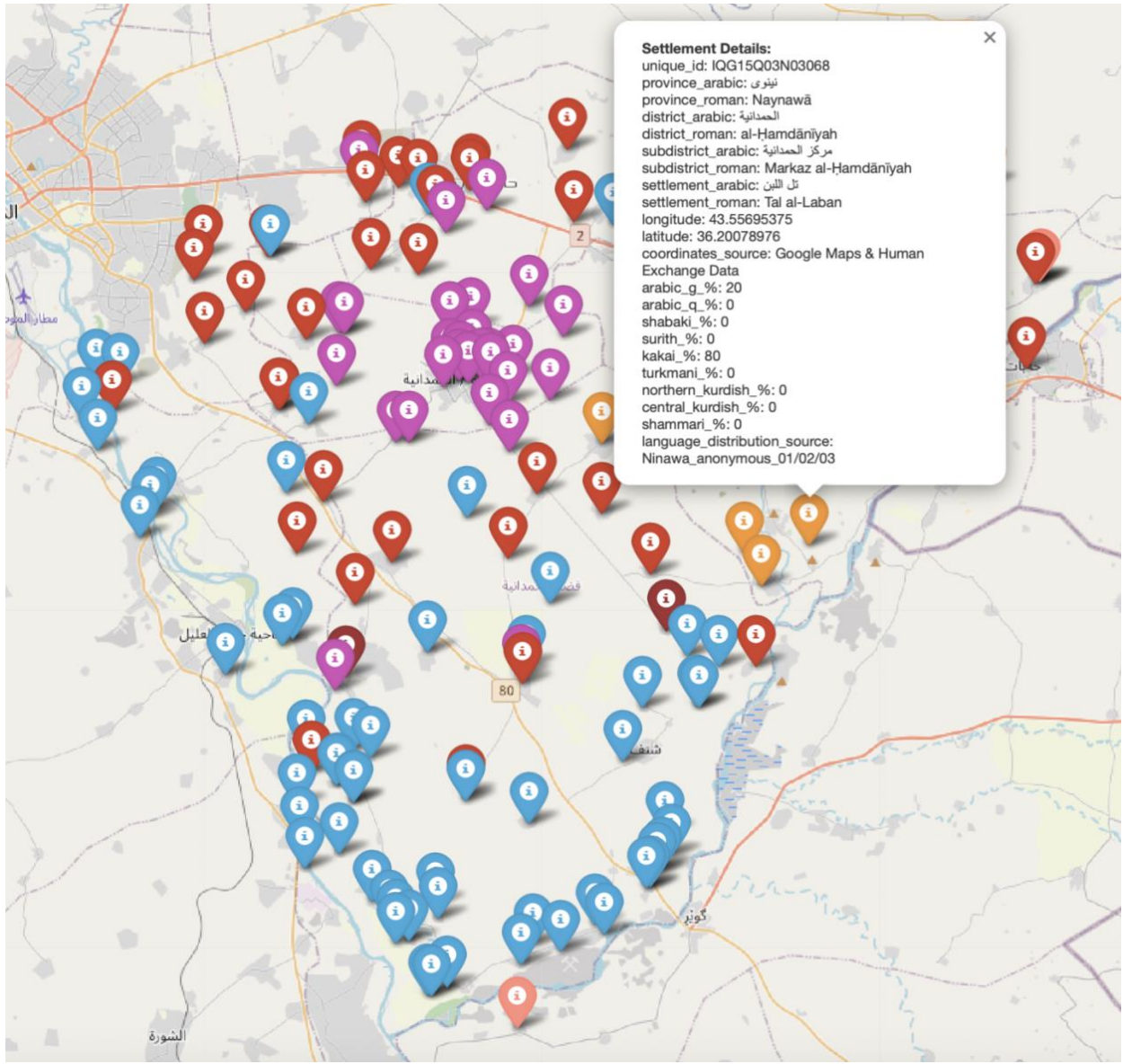

Map 14 – this map shows the settlements in pined in different colors based on the language spoken in them. Once we click on the pin, we can see all the hover data we choose to include such as place names in English and Arabic, language distribution data, and other location information. It can be noted here that there is no legend to explain which language is represented by each of the colors. This was one of the limitations of the AI models, not being able to provide a legend within the map even after addressing that the problem, the models gave a rectified code that still did not work.

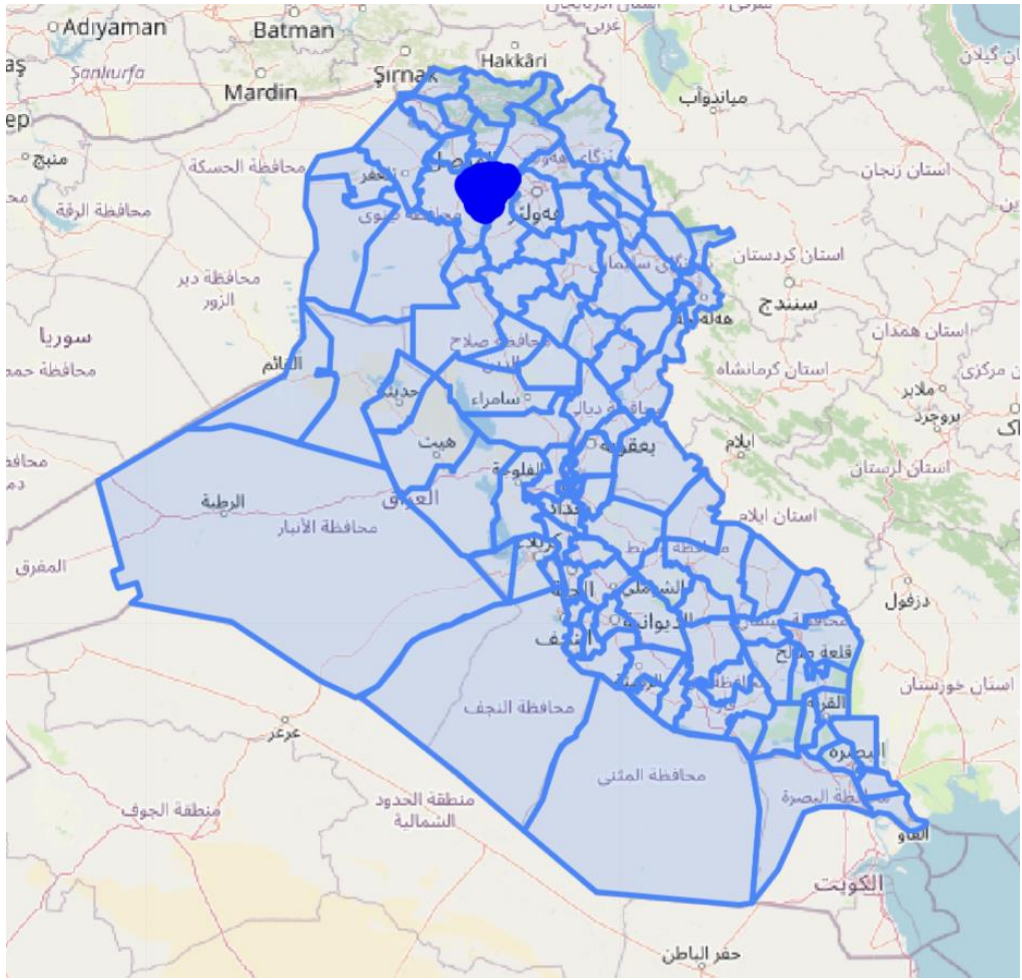

Map 15 – This map shows how the GeoJSON shape file is overlaid on the base maps, here Open Street Map, to show the administrative borders because it was difficult to see the borders on the base map. This map is interactive, which means that we can zoom it in so we can see the plotted data as points on the settlements. Once we click on a point, we can see the hover data related to it.

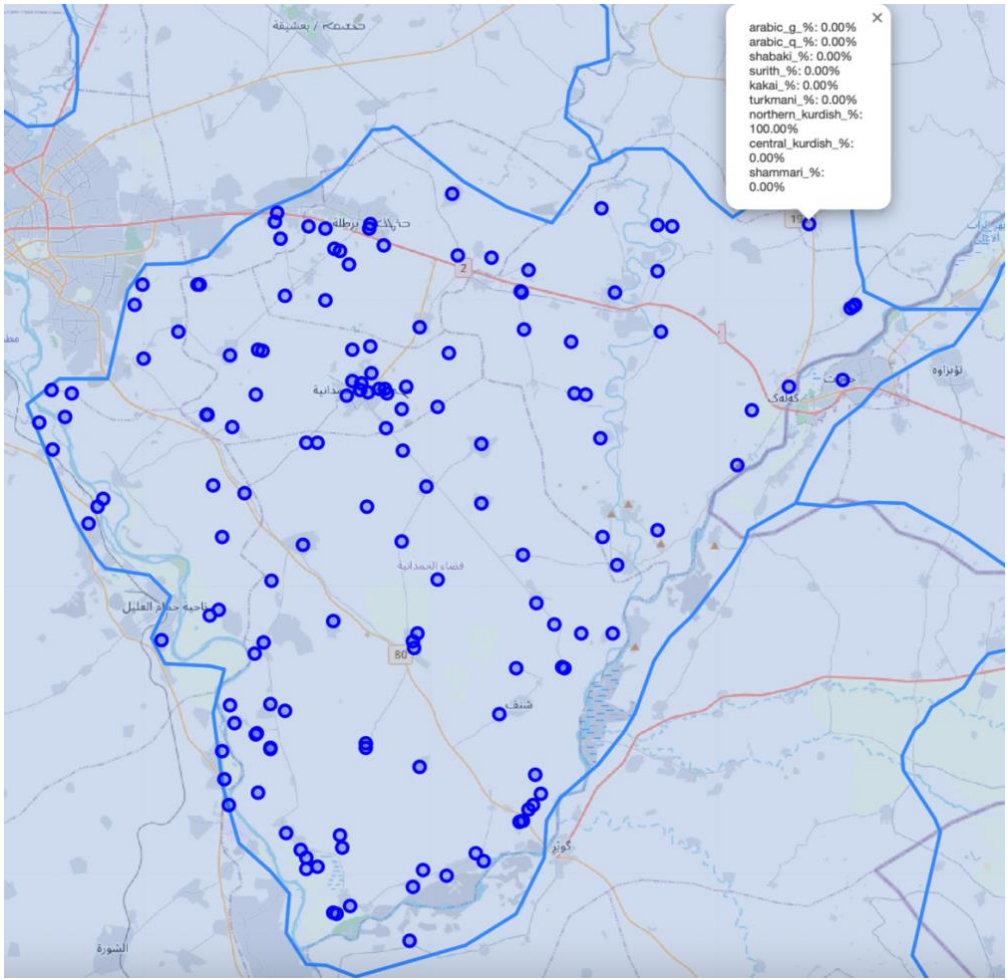

Map 16 – This is a zoomed version of the map above (Map 15), it can be seen that after zooming the map, we can see the points on the settlements and the hover data on one of them.

**Appendix II**

**General information**

*(to be filled out by the field researcher <u>before</u> or <u>after</u> fieldwork)*

Name of person filling out this questionnaire: \_\_\_\_\_\_\_\_\_\_\_\_\_\_\_\_\_\_\_\_\_\_\_\_\_

Email address: \_\_\_\_\_\_\_\_\_\_\_\_\_\_\_\_\_\_\_\_\_\_\_\_\_

Location where the questionnaire was filled out: \_\_\_\_\_\_\_\_\_\_\_\_\_\_\_\_\_\_\_\_\_\_\_\_\_

Date filled out: \_\_\_\_\_\_\_\_\_\_\_\_\_\_\_\_\_\_\_\_\_\_\_\_\_

Name of person(s) checking / analyzing the questionnaire data (checker to fill this in): \_\_\_\_\_\_\_\_\_\_\_\_\_\_\_\_\_\_\_\_\_\_\_\_\_

Date checked / analyzed (checker to fill this in): \_\_\_\_\_\_\_\_\_\_\_\_\_\_\_\_\_\_\_\_\_\_\_\_\_

Date approved for upload (editors to fill this in): \_\_\_\_\_\_\_\_\_\_\_\_\_\_\_\_\_\_\_\_\_\_\_\_\_

Language variety (language, dialect, subdialect, etc.): \_\_\_\_\_\_\_\_\_\_\_\_\_\_\_\_\_\_\_\_\_\_\_\_\_

Language data from which province and which exact village(s)/town(s) \_\_\_\_\_\_\_\_\_\_\_\_\_\_\_\_\_\_\_\_\_\_\_\_\_

Is the data from your own fieldwork and/or from a published source? \_\_\_\_\_\_\_\_\_\_\_\_\_\_\_\_\_\_\_\_\_\_\_\_\_

*If the data is from your own existing field notes, please ensure that it is from a <u>single</u> settlement. Then fill in the following details:*

**Permission text and speaker details**

*My name is .... I come from .... I am a linguist from the University of .... I want to study and learn about the languages and dialects of* [region] *and help preserve knowledge about these languages for the future. I use the things that I learn for research and eventual publication in print, online in the Atlas of the Languages of Iraq, and in an online language archive. Do you give me the permission to ask you questions about your language, and to record and freely use what I learn?*

    [make sure the answer is loud and clear enough to be audible on the recording]

**If the speakers have given their consent, say:**

*If you wish to discontinue this interview at any point or withdraw your consent for me to use anything that you've shared, including audio or video recordings, you are free to do so. I am able to withdraw your contributions until the point when they are published or archived.*

Then, you can continue with the following questions, and the rest of the interview:

Would you like us to write your name(s) down, or would you like to remain anonymous?
\_\_\_\_\_\_\_\_\_\_\_\_\_\_\_\_\_\_\_\_\_\_\_\_\_\_\_\_\_

Name of speaker(s) (only put this if speakers have requested that their names be used):

(speaker 1)    ______________________________
(speaker 2)    ______________________________
(speaker 3)    ______________________________

Other details about the speakers (include this information for all speakers):

    Age:    Gender:  Profession:

(speaker 1)    ________    ________    ____________________
(speaker 2)    ________    ________    ____________________
(speaker 3)    ________    ________    ____________________

Languages spoken, along with:

    a) speaker's <u>L1</u> (first language learned from parents in the home), or <u>L2+</u> (an additional language)?

    b) speaker's stated degree of <u>oral</u> proficiency in each language (*full = 5 / good = 4 / some = 3 / little = 2 / very little = 1*)

    c) frequency of usage in the past year (*every day / at least once every week / at least once every month / at least once in the past year / never in the past year*)

    Language:    L1 or L2+?    Proficiency:    Frequency:

(speaker 1)    ___________  _______    ________    _____________
                  ___________  _______    ________    _____________
                  ___________  _______    ________    _____________
                  ___________  _______    ________    _____________
                  ___________  _______    ________    _____________

(speaker 2)    ___________  _______    ________    _____________
                  ___________  _______    ________    _____________
                  ___________  _______    ________    _____________
                  ___________  _______    ________    _____________
                  ___________  _______    ________    _____________
                  ___________  _______    ________    _____________

(speaker 3)    ___________  _______    ________    _____________
                  ___________  _______    ________    _____________
                  ___________  _______    ________    _____________
                  ___________  _______    ________    _____________
                  ___________  _______    ________    _____________
                  ___________  _______    ________    _____________

Languages the speaker can read and write, and the speaker's stated degree of <u>written</u> proficiency in each language (full = 5 / good = 4 / some = 3 / little = 2 / very little = 1)

(speaker 1)    ______________________________________________________

(speaker 2)     \_\_\_\_\_\_\_\_\_\_\_\_\_\_\_\_\_\_\_\_\_\_\_\_\_\_\_\_\_\_\_\_\_\_\_\_\_\_\_\_\_\_\_\_\_\_\_\_

(speaker 3)     \_\_\_\_\_\_\_\_\_\_\_\_\_\_\_\_\_\_\_\_\_\_\_\_\_\_\_\_\_\_\_\_\_\_\_\_\_\_\_\_\_\_\_\_\_\_\_\_

Level of education (none / elementary / secondary / post-secondary):

(speaker 1)     \_\_\_\_\_\_\_\_\_\_\_\_\_\_\_\_\_\_\_\_\_\_\_\_\_\_\_\_\_\_\_\_\_\_\_\_\_\_\_\_\_\_\_\_\_\_\_\_

(speaker 2)     \_\_\_\_\_\_\_\_\_\_\_\_\_\_\_\_\_\_\_\_\_\_\_\_\_\_\_\_\_\_\_\_\_\_\_\_\_\_\_\_\_\_\_\_\_\_\_\_

(speaker 3)     \_\_\_\_\_\_\_\_\_\_\_\_\_\_\_\_\_\_\_\_\_\_\_\_\_\_\_\_\_\_\_\_\_\_\_\_\_\_\_\_\_\_\_\_\_\_\_\_

**Sociolinguistic information**

The following portion of the questionnaire, which deals with language use in the context of a single settlement, is adapted from [Anonby & Yousefian's (2011) sociolinguistic study](#).
*Please answer the questions as you are able, providing estimates if necessary. You may mark any item for which you are unable to provide an answer, with "?".*

1. What is the name of your community (city, town, village, etc.)? ______________________________

2. What languages are spoken in the community as a mother tongue (that is, the first language a person learns at home as a child)?
   ____________________________________________________________________

3. What proportion (approximate percentage) of the population in the community speaks each of these languages as a mother tongue?
   ____________________________________________________________________

4. What do the people in the community call their own language(s) [the most general language grouping possible] and dialect(s) [more local groupings at the regional and local level, i.e., their particular variety of the language they mention] in their own language(s)?
   ____________________________________________________________________

5. What do the people in the community call their own ethnic group(s) in their own language(s)? [Ask the speaker to give ethnic group names at all relevant levels, from most general to most specific, as far as they are known, and specify each level in parentheses beside each of the ethnic names.
   ____________________________________________________________________

6. What is the name of the community (city, town, village, etc.) in each of the languages that are spoken there?
   ____________________________________________________________________

7. What proportion of the population (circle one answer for each):

speaks Arabic fluently?   *all     most     some     a few     none*
speaks some Arabic?       *all     most     some     a few     none*
speaks no Arabic?         *all     most     some     a few     none*

8. Besides their mother tongue(s) and Arabic, what additional languages and dialects do the people of the community speak?
   ____________________________________________________________________

9. What proportion (*all/most/some/a few/none*) of the children in the community are not learning the mother tongue of their parents as their mother tongue? __________

10. Which language communities do these children come from? (At the same time, specify the proportion of children – *all/most/some/a few/none* – who are not learning the mother tongue of their parents, for each language community.)
    ____________________________________________________________________

11. What language(s) are these children now learning as a mother tongue?
___________________________________________________________